\documentclass[letterpaper, 10 pt, conference]{ieeeconf}  %
\pdfoutput=1                                              %

\IEEEoverridecommandlockouts                              %
\newcommand{\fref}[1]{Fig.~\ref{#1}}
\newcommand{\tref}[1]{Table~\ref{#1}}

\let\labelindent\relax
\usepackage[inline]{enumitem}
\usepackage{booktabs}	%
\usepackage{graphics} %
\usepackage{epstopdf}	%
\usepackage{epsfig} %
\usepackage[utf8]{inputenc} 	%
\usepackage{amsmath} %
\usepackage{amssymb}  %
\usepackage{cite}
\usepackage[colorinlistoftodos,prependcaption]{todonotes}
\usepackage{subcaption}
\usepackage{hyperref}

\title{\LARGE \bf
Robust Dense Mapping for Large-Scale Dynamic Environments
}

\author{Ioan Andrei B\^{a}rsan$^{1,2,3}$, Peidong Liu$^{1}$, Marc Pollefeys$^{1,4}$ and Andreas Geiger$^{1,5}$%
\thanks{$^{1}$Computer Vision and Geometry Group, ETH Z\"{u}rich, Switzerland
  \texttt{\scriptsize\{peidong.liu, andreas.geiger, marc.pollefeys\}@inf.ethz.ch}}%
\thanks{$^{2}$University of Toronto, Canada \texttt{\scriptsize iab@cs.toronto.edu}}
\thanks{$^{3}$Uber ATG Toronto, Canada}
\thanks{$^{4}$Microsoft, Redmond, USA}%
\thanks{$^{5}$Autonomous Vision Group, Max Planck Institute for Intelligent Systems, T\"{u}bingen, Germany}%
}

\begin{document}
\maketitle
\thispagestyle{empty}
\pagestyle{empty}

\begin{abstract}
We present a stereo-based dense mapping algorithm for large-scale dynamic urban environments.
In contrast to other existing methods, we simultaneously reconstruct the static background, the moving objects, and the potentially moving but currently stationary objects separately, 
which is desirable for high-level mobile robotic tasks such as path planning in
crowded environments. 
We use both instance-aware semantic segmentation and sparse scene flow to 
classify objects as either background, moving, or potentially moving, 
thereby ensuring that the system
is able to model objects with the potential to transition from static to dynamic, such
as parked cars.
Given camera poses estimated from visual odometry, both the background and 
the (potentially) moving objects are reconstructed separately by fusing the 
depth maps computed from the stereo input. 
In addition to visual odometry, sparse scene flow is also used to estimate the 
3D motions of the detected moving objects, in order to reconstruct them accurately.
A map pruning technique is further developed to improve reconstruction accuracy
and reduce memory consumption, leading to increased scalability.
We evaluate our system thoroughly on the well-known KITTI dataset.
Our system is capable of running on a PC at approximately 2.5Hz, 
with the primary bottleneck being the instance-aware semantic segmentation, which is a limitation we hope to address in future work.
The source code is available from the project 
website\footnote{\url{http://andreibarsan.github.io/dynslam}}.
\end{abstract}

\section{INTRODUCTION}
Over the course of the past two decades, the field of 3D visual perception has undergone
a dramatic evolution, enabling many new applications such as self-driving cars, automatic
environment mapping, and high-quality object scanning using consumer-grade sensors such as
the Microsoft Kinect.
Nevertheless, despite the field’s rapid evolution, numerous open problems still remain.
Among them is the task of real-time
large-scale mapping for dynamic outdoor environments.

\begin{figure}[htb]
	\centering
    \includegraphics[width=0.45\textwidth]{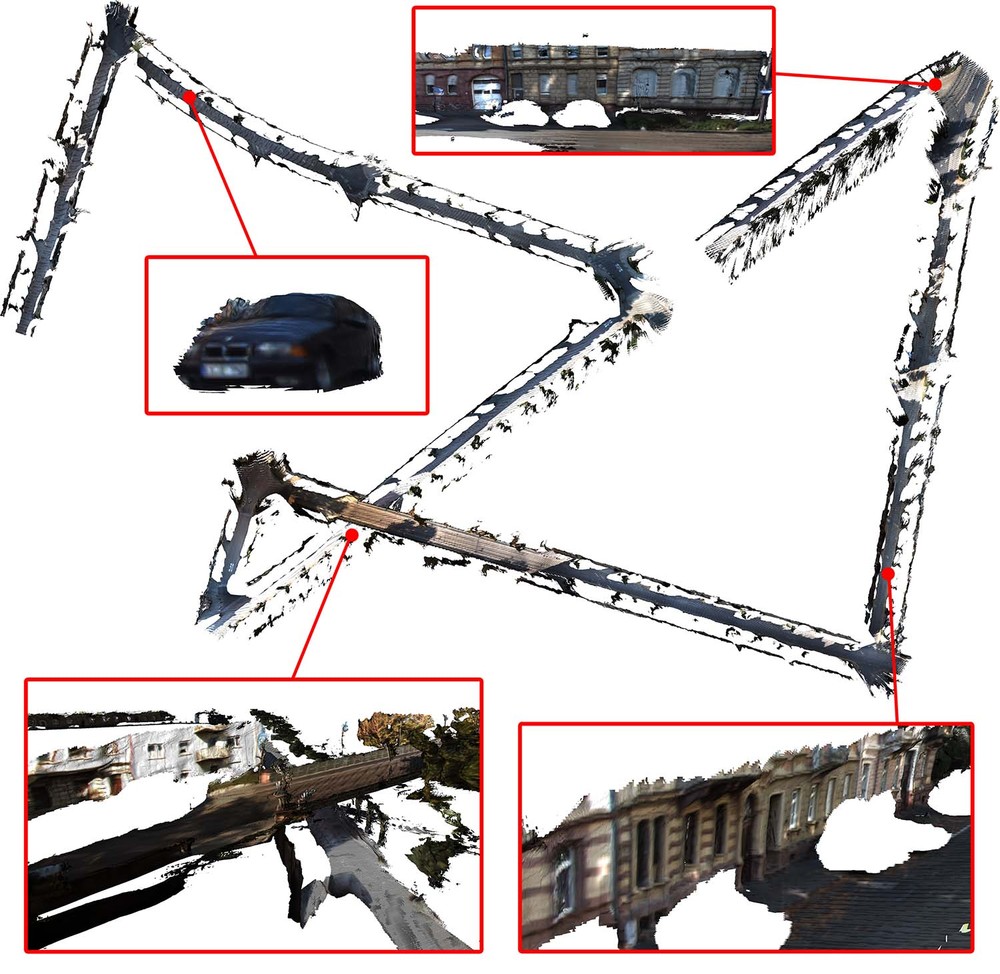}
    \caption{A static map with all moving and potentially moving objects removed, as well
    as the reconstruction of one independently moving vehicle encountered
    in the environment.\label{fig:intro}}
\end{figure}

Map representations range from the conventional
point cloud~\cite{Scaramuzza_ICRA_2014,Furukawa_PMVS_PAMI_2010,mur2015orb}
and mesh-based~\cite{Whelan2012} approaches to more sophisticated ones based on, e.g.,
volumetric methods~\cite{schoeps_3dv_2015,oleynikova_ICRA_2017,Shen_ICRA_2017}.
While point-based approaches can provide accurate reconstructions, the level of detail depends on the number of points used, which can become prohibitive for real-time applications.
Furthermore, without additional meshing steps, sparse representations do not easily accommodate tasks like path
planning, which require line-of-sight computations.
In contrast, volumetric representations can be used directly by path planning.
Such methods represent 3D geometry as a regularly sampled 3D grid by means of an implicit function which either models occupancy or encodes the distance to the closest surface.
Although memory usage is one of the main limitations of these methods, multiple techniques have been proposed for dealing with
this problem using either hashing \cite{Niessner2013b}, octrees \cite{Steinbruecker_ICRA_2014} or data-adaptive discretization \cite{Labatut_ICCV_2007}.

The sensors typically used in 3D mapping include depth~\cite{Niessner2013b},
monocular~\cite{Shen_ICRA_2017}, and stereo~\cite{Reddy2015} cameras.
While depth cameras can readily provide high-quality depth measurements, their usage is
limited to indoor environments.

On the other hand, conventional stereo and monocular cameras can operate
successfully in both indoor and outdoor environments, under a wide range of
lighting conditions.

For real-world applications, mobile robots usually need to operate in large-scale,
complex environments, e.g., autonomous driving in a crowded city, which is full of pedestrians and moving vehicles.
These scenarios pose a challenge for most existing 3D mapping algorithms, which usually
assume that the surrounding environments are static, meaning that dynamic objects would
corrupt the reconstructed map if not explicitly considered.
While numerous existing mapping systems are capable of operating within dynamic
environments, they typically achieve this robustness by treating sensory input associated
with dynamic objects as noise, and ignoring it in order to preserve the consistency of the static map~\cite{Vu2007,Reddy2015,Vineet2015}.
However, this process discards valuable information about the dynamic objects in the environment, which can potentially be used in applications such as path planning and traffic
understanding.

In this paper, we propose a robust dense mapping algorithm for large scale dynamic
environments.
In particular, we use a volumetric representation
based on voxel block hashing~\cite{Niessner2013b,Kaehler2015} for large scale environments.
A pair of stereo cameras is used to infer the egomotion and reconstruct the surrounding
world. Semantic cues and the sparse scene flow are used to classify objects as either
static background, moving, or potentially moving but currently stationary.
The objects are then treated separately such that a high-quality corruption-free dense map of the static background can be inferred.
Instead of discarding dynamic objects, we reconstruct them separately, which has potential for future applications such as high-level robotic decision making.
Each object is represented by a separate volumetric map and its motion is tracked.
We focus our attention on reconstructing rigid objects, such as cars and trucks,
since reconstructing articulate objects (e.g., pedestrians) from multiple
observations is more computationally demanding~\cite{Newcombe2015}.
To reduce the overall memory consumption and increase the accuracy of the resulting
reconstructions, a map pruning technique is proposed.
\fref{fig:intro} shows a reconstructed static map with (potentially) moving objects removed,
and the reconstruction of an independently moving vehicle encountered in the environment.

The main contributions of our work are summarized as follows.
\begin{enumerate*}
  \item We develop an efficient stereo-based dense mapping algorithm robust to
    dynamic environments.
  \item The system builds a high-quality static map and individual 3D
    reconstructions of moving and potentially moving objects in an online
    manner.
  \item A map pruning technique is proposed to further improve the mapping
    accuracy and reduce memory consumption, thereby increasing the system's
    scalability.
\end{enumerate*}

\section{RELATED WORK}
There is a large body of recent and ongoing research in fields related to autonomous robotics,
such as dense mapping, object tracking, and semantic segmentation.
In this section, we focus on the parts of this research most relevant to our work.

Reddy et al.~\cite{Reddy2015} describe a stereo-based algorithm for robust SLAM that works in both static and dynamic environments. They separate the static background from the moving objects by minimizing a joint CRF-based semantic motion segmentation energy function.
Both the ego-motion and the geometric structure are optimized using bundle adjustment by incorporating additional geometric and semantic priors. The experimental results are good but not without
limitations. The reconstructed moving objects are not consistent with the real objects due to noisy depth estimation. The system is not applicable to large scale environments and is not able to run in (near) real-time due to the expensive optimization of the aforementioned energy function.

Vineet et al.\ propose a large scale semantic dense reconstruction algorithm in \cite{Vineet2015}.
The system runs in near real time, but moving objects are not considered explicitly, being
reconstructed together with the static background.
To avoid map corruption, they propose an object class-dependent weighting scheme for the depth
map fusion.
In particular, areas of the map corresponding to moving objects are updated using larger
weights, i.e., more aggressively, in order to ensure that they remain consistent.

Kochanov et al.~\cite{Kochanov2016} propose a large scale dense mapping system for dynamic environments. They use stereo cameras to build a dense semantic map of an environment which also incorporates information on the dynamics of the scene. Their method is centered around augmenting a sparse voxel grid to also include a running estimate of each cell's scene flow, in addition to the semantic and occupancy information. However, the described pipeline does not run in real time. Among its components, it leverages the dense scene flow method of Vogel et al.~\cite{vogel2013piecewise}, which can take up to 300 seconds for a single frame. The system is also limited to detecting and tracking dynamic objects, without attempting to reconstruct them.

Jiang et al.~\cite{Jiang2016} present a SLAM system for dynamic environments which uses LIDAR and cameras to reconstruct both the static map, as well as the dynamic objects within, using 3D sparse subspace clustering of key-point trajectories to identify and segment dynamic objects. Nevertheless, the presented system does not run in real time, requiring roughly 12 minutes to process a 70-frame (i.e., seven-second) sequence.
A follow-up work from the same authors~\cite{Jiang_HAL_2017} improves the
system's robustness to challenges such as partial occlusions by using
3D flow field analysis.

Co-Fusion~\cite{Ruenz2017} is another real-time dense mapping system capable of reconstructing both a static background map, as well as dynamic objects from a scene. The paper presents a real-time pipeline which can segment and track dynamic objects based on either motion or semantic cues, reconstructing them separately using a surfel-based representation.
While most of the experiments are performed on indoor RGB-D sequences with modest camera motion
(both real-world and synthetic), some results on the Virtual KITTI dataset~\cite{gaidon2016virtual} are also presented. However, both the quality and the scale of the reconstructions performed on the Virtual KITTI data are
limited, in comparison to their indoor sequences. Moreover, the authors do not address the problem of noise reduction for 3D reconstruction, which is an important consideration for stereo-based large-scale dense mapping of outdoor environments.

\section{DYNAMIC RECONSTRUCTION}

\begin{figure*}
  \centering
  \includegraphics[width=0.75\textwidth]{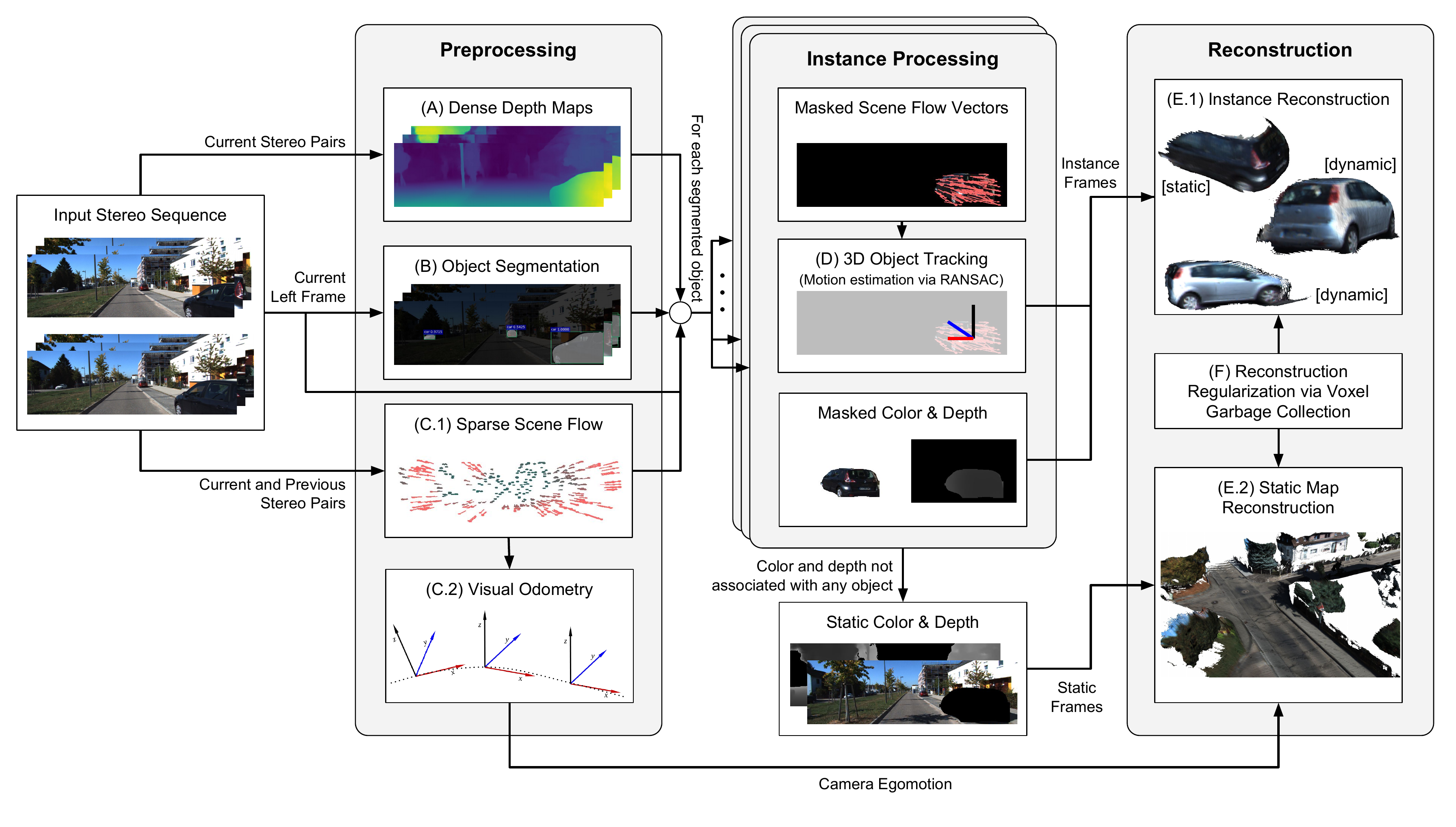}
  \caption{An overview of the system pipeline.}
  \label{fig:pipeline}
\end{figure*}

\fref{fig:pipeline} shows an overview of our dense mapping system. 
At a glance, the system performs the following steps at every frame:
\begin{enumerate}
\item Pre-process the input by computing a dense depth map and sparse 
  scene flow from the stereo pair, as well as an instance-aware semantic
  segmentation of the RGB data from the left camera. 
  (Steps (A), (B), and (C.1) from \fref{fig:pipeline}.)
\item Compute VO from the sparse scene flow (Step (C.2)).
\item Separate input (color, depth, and sparse flow) into multiple
  ``virtual frames'': one for the background, plus one for each potentially
  dynamic object in the frame.
\item Estimate the 3D motion of each new detection using the scene flow
  and semantic segmentation information, comparing it to the camera egomotion
  to classify each object as static, dynamic, or uncertain (Step (D)).
\item For each rigid object of interest (moving or potentially moving),
	initialize or update its reconstruction (Step (E.1)).
\item Update the static map reconstruction (Step (E.2)).
\item Perform voxel garbage collection to remove voxels allocated spuriously
	due to artifacts in the depth map (Step (F)).
\end{enumerate}
We will now describe the details of each component of the pipeline.

\subsection{Dense depth map computation}
The pipeline begins with several preprocessing steps, the first of which is
estimating a dense depth map from the input stereo image pair.  In this work,
we investigate two different state-of-the-art real-time stereo matching
techniques, Efficient Large-scale Stereo Matching
(ELAS)~\cite{Geiger_ACCV_2010} and DispNet~\cite{Mayer_CVPR_2016}.
The former method is geometry-based, relying on a sparse keypoint matching
phase, followed by a dense matching one, whereas the latter is an approach
based on deep learning consisting of a neural network trained to directly
regress disparity maps from stereo inputs.  We have chosen to evaluate our
system using disparity maps produced by both methods given their highly
distinct approaches, which exhibit different strengths and weaknesses: ELAS
depth maps tend to be sharper but incomplete, with many gaps present in
reflective and transparent areas, while the results of DispNet are denser and
robust to non-lambertian surfaces, but less sharp, especially around object
boundaries.

\subsection{Object segmentation and 2D tracking}
An instance-aware semantic segmentation algorithm is used to recognize dynamic and 
potentially dynamic objects from a single image (e.g., both actively moving and parked cars). 
We use Multi-task Network Cascades~\cite{Dai_CVPR_2016} (MNC),
a state-of-the-art deep neural network architecture for this task.
This component detects and classifies object instances in an input image using
the 20 labels from the Pascal VOC2012 dataset.

The object detections are computed independently in every frame.
An additional inter-frame association step is therefore needed
in order to consistently track the objects across multiple frames, a
prerequisite for reconstruction.
We achieve this by greedily associating new detections with existing 
tracks by ranking them based on the Intersection-over-Union 
(IoU) score between a new detection and the most recent frame in a track.
New detections which are not matched to any existing track are used 
to initialize new tracks.
The system only tracks rigid objects which can be reconstructed further
in the pipeline. Information associated with (potentially) dynamic 
non-rigid objects, such as pedestrians, is removed from the frame
to prevent it from interfering with the background reconstruction, but does
not get passed to the 3D motion estimation and reconstruction components.

\subsection{Sparse scene flow and visual odometry}
Both the sparse scene flow and the vehicle egomotion are computed using 
\texttt{libviso2}~\cite{Geiger2011}. Following the method described in~\cite{Geiger2011}, the estimation of the sparse scene flow is based on two-view and temporal stereo. 
More specifically, simple blob and corner features are matched between the current left and right frames, and the previous ones, resulting in four-way matches.
These matches are equivalent to pairs of 3D points from consecutive time steps, 
i.e., the scene flow. The matches are used in the computation of the visual odometry, 
as well as in estimating the 3D motion of the tracked objects.
A RANSAC-based approach is used to compute visual odometry from the sparse scene flow vectors. At every frame, the six-degree-of-freedom pose of the stereo camera relative to the previous frame is estimated by minimizing the reprojection error of the 3D points triangulated from the feature positions.

\subsection{3D object tracking}
3D object tracking occurs on a frame-to-frame basis and uses the
sparse scene flow computation described in the previous pipeline stage as input.
The masked scene flow associated with a specific object instance is used as input to estimate the motion of a virtual camera with respect to the object instance, which is assumed to be static.
If the estimation is successful, then the 3D motion of the object is equal to the inverse of the virtual camera’s motion. For static objects, this obviously means that the resulting 3D object motion will be nearly identical to the camera’s egomotion. 
This can be used to classify objects with known motion as either static or dynamic.
The motion of the virtual camera is computed the same way as the egomotion,
using the robust approach from \texttt{libviso2} described in the previous section.

\subsection{Static map and individual object reconstructions}
Our system uses InfiniTAM~\cite{Kaehler2015} for volumetric fusion. 
InfiniTAM is an efficient framework for real-time, large scale depth fusion and tracking. Instead of fusing depth maps into a single volumetric model, we separate the static background from the dynamic objects. 
The vehicle egomotion computed by the visual odometry is used to fuse the static parts 
of the input color and depth maps, which are identified based on the output of the
instance-aware semantic segmentation component. 
Both moving and potentially moving objects are reconstructed individually. 
Each object is represented by a single volumetric model, whose coordinate system is centered
around the pose of the camera in the frame where the object was first observed. 
The estimated 3D motions of the individual objects are used for the object
volumetric fusion. 
For every relevant object detection at a given time step $t$, its corresponding RGB and depth data are
extracted using the mask resulting from the segmentation procedure, resulting
in a ``virtual frame'' containing only that particular object's data. This
virtual frame, together with the object's estimated 3D motion, represents the
input to the object volumetric reconstruction.

\subsection{Fixed-lag map pruning}
\begin{figure}
	\centering
	\includegraphics[width=0.45\textwidth]{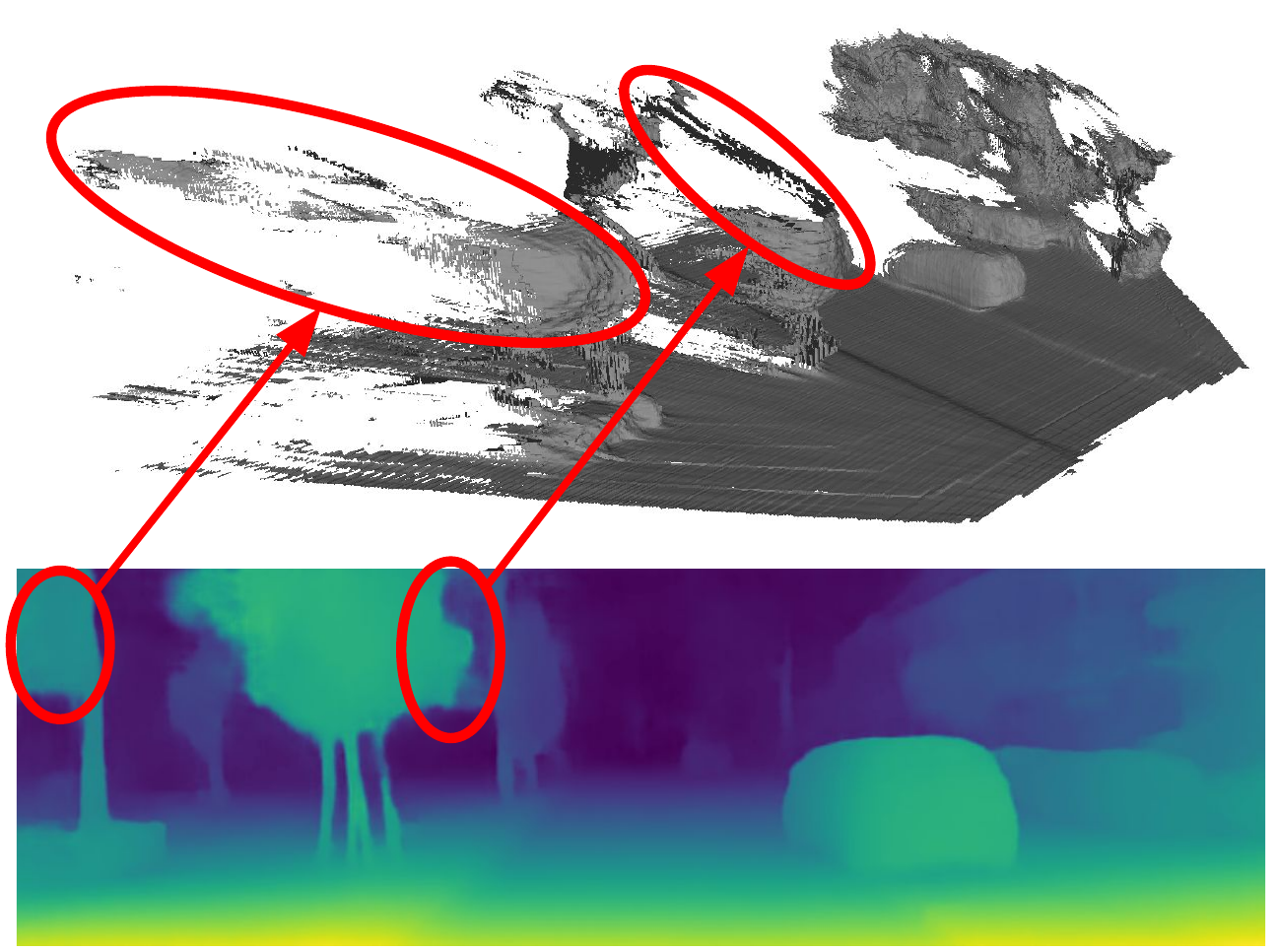}
    \caption{Examples of the streak-like artifacts produced in 3D reconstructions
    by noisy depth maps.\label{fig:streaks}}
\end{figure}

When performing volumetric reconstruction at a small scale, such as when scanning
room-sized environments with an RGBD camera, the magnitude of the noise associated with
the depth estimation is relatively small, compared to the error associated with estimating
depth from stereo~\cite{Lenz2011,Giovani2015b}. 
Therefore, the development of methods for actively tackling noise-induced
artifacts is not a primary consideration when dealing with small-scale reconstructions, 
but it becomes a necessity when switching to outdoor stereo reconstruction. 
\fref{fig:streaks} shows an example of the artifacts produced when rendering a stereo
depth map in 3D. The outlines of most objects have a comet-like trail facing away from
the direction from which they were perceived. The streaks not only reduce the map quality,
but also lead to increased memory usage, since they occupy large volumes,
negatively impacting the system's scalability.

In order to reduce the impact of the noise, and improve the reconstruction quality
while reducing the memory footprint of our system, we turn to a technique first proposed by
Nie{\ss}ner et al.~\cite{Niessner2013b}. Their work describes a garbage collection
method for removing voxel blocks allocated due to noise and moving objects.
For every $8 \times 8 \times 8$ voxel block, the minimum absolute value of
the truncated signed distance function (TSDF) is computed together with the maximum
measurement weight. 
Blocks with a minimum absolute TSDF above a certain threshold, which correspond to
areas far enough from the represented surface so as not to contribute meaningfully to it,
and blocks with a maximum weight of zero, which are  blocks not containing any 
measurement information, are removed from the reconstruction, reducing artifacts and
freeing up memory.

We extend this method in two ways: First, we increase its granularity, making it
capable of operating on a per-voxel basis as opposed to a per-voxel block basis, 
meaning that noisy voxels can be deleted even when the block to which they belong cannot be. Note that memory is still only freed when entire voxel blocks are deleted. This is due to 
the voxel block hashing-based map representation used in InfiniTAM, which does not 
allow individual voxels to be deallocated. 

Second, we improve the method's scalability by allowing it to take advantage of
InfiniTAM's visible block lists, removing the need to run the method on the entire
map at every frame, which would have prohibitively high computational costs.
This improvement is based on two insights: (1) The movement of the vehicle-mounted
camera is not ``loopy,'' as is common in indoor sequences, with loop closures 
being rare. (2) Cleaned-up areas do not need to be re-processed until they are
revisited by the camera. 

InfiniTAM keeps track of a list of currently visible blocks, for efficiency reasons.
By storing a snapshot of this list at every time step, the history of which blocks 
were visible over time can be recorded. 
At every time step $t$, we process the voxel blocks visible
at time $t - k_\text{minAge}$, where $k_\text{minAge}$ is the minimum age of a
voxel block in terms of time steps, which is set empirically such that the majority
of the voxel blocks processed by the collection are no longer visible\footnote{In most of our experiments it is set to 300, corresponding to 30 seconds of driving time.}.
In other words, the garbage collection is performed in lockstep with the fusion:
after processing the $n$th input frame the system performs voxel garbage collection 
on the blocks visible at the time of input frame $n - k_\text{minAge}$.

\begin{figure*}
	\centering
    \begin{subfigure}[t]{0.38\textwidth}
    	\centering
        \includegraphics[width=\linewidth]{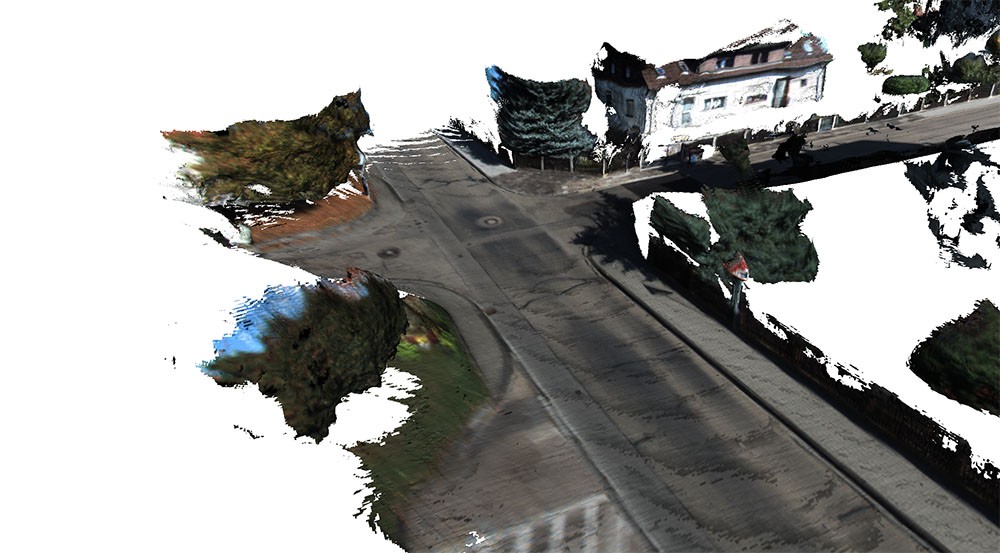}
        \caption{Voxel GC with the noise threshold $\Delta_\text{weight} = 6$.}
    \end{subfigure}~
    \begin{subfigure}[t]{0.38\textwidth}
    	\centering
        \includegraphics[width=\linewidth]{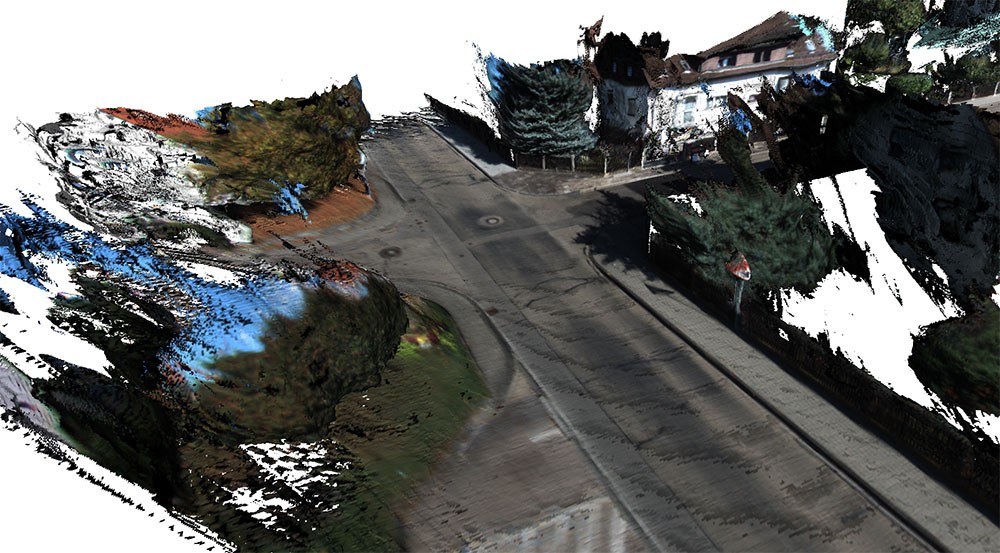}
        \caption{No voxel GC.}
    \end{subfigure}~%
    \caption{Comparison of the same scene reconstructed with voxel garbage collection and without it.\label{fig:gc-qualitative}}
\end{figure*}

In scenes where the static map can easily occupy hundreds of thousands of voxel blocks, 
the visible list at every given frame typically holds around 8,000--12,000 blocks, resulting 
in a 10-fold reduction of computational costs, as compared to running the process over the 
entire volume at every step. 
On an nVIDIA GeForce GTX 1080 GPU, this results in an overhead of less than 1ms at every frame. \fref{fig:gc-qualitative} shows the impact of the voxel
garbage collection on a scene reconstruction.
Note that we only apply this technique when performing voxel garbage collection on the
static map. Given their limited dimensions, car reconstructions are always pruned
in their entirety at every frame.

\section{EXPERIMENTS}
We use the well-known KITTI Vision Benchmark Suite~\cite{Geiger2012} to evaluate
our system.
Both qualitative and quantitative evaluations of the accuracy of the obtained 3D
reconstructions are performed.
The effectiveness of the fixed-lag map regularizer is also evaluated.
Note that we focus our attention on the mapping aspect, and do not evaluate
the localization accuracy, as it is not the primary focus of this paper.

\subsection{Qualitative accuracy evaluation}
\begin{figure*}
  \centering
  \begin{subfigure}[t]{0.45\textwidth}
    \centering
    \includegraphics[width=\linewidth]{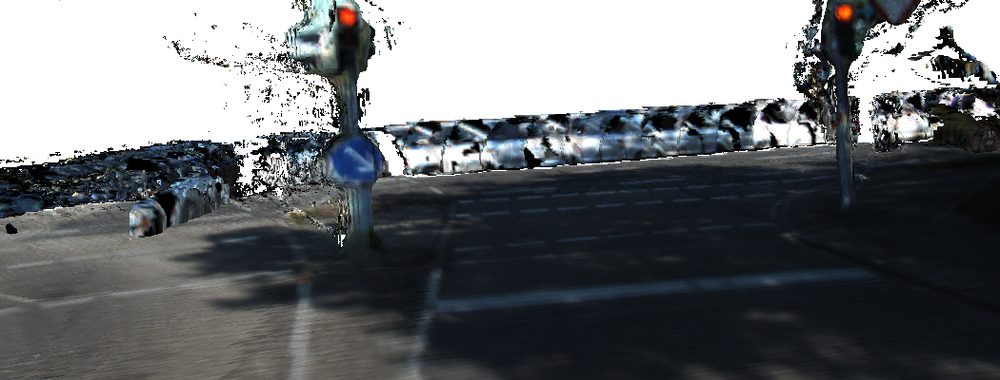}
    \caption{Static fusion is prone to corrupt the environment map with streaks
    and other artifacts produced by independently moving objects.}
  \end{subfigure}
  ~
  \begin{subfigure}[t]{0.45\textwidth}
    \centering
    \includegraphics[width=\linewidth]{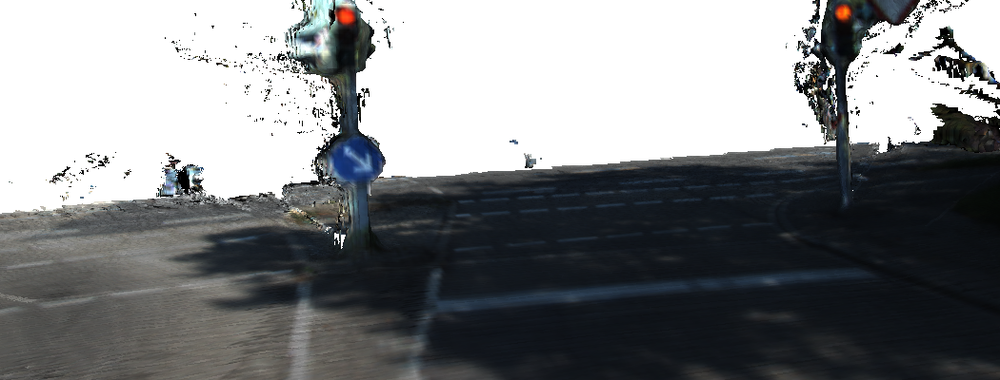}
    \caption{Dynamic fusion, the primary operating mode of our system, prevents
    vehicle trails and leftover halos from being integrated into the map.}
  \end{subfigure}
  \caption{Reconstructions produced by Dynamic and Standard fusion on KITTI
  tracking sequence 01.\label{fig:ghost-cars}}
\end{figure*}
\fref{fig:ghost-cars} shows two example reconstructions of the same view by dynamic-aware fusion and standard fusion on a KITTI sequence. It shows that individually moving vehicles corrupt the environment map when ignored,
while the dynamic-aware fusion provides a corruption-free environment map.

\subsection{Quantitative accuracy evaluation}
We base our experiments on the video sequences from (a) the KITTI odometry and (b) the KITTI tracking benchmarks, using the LIDAR as a ground truth for evaluating the quality of both the reconstructed static maps, as well as of the dynamic object instances. Similar to the works of Sengupta et al.~\cite{Sengupta2013d} and
Vineet et al.~\cite{Vineet2015},
we compare the LIDAR ground truth to the current reconstruction and to the
input depth map at every frame by projecting everything to the left camera's frame. This follows the same principle as the
2015 KITTI stereo benchmark with the main difference being that we average the accuracy scores across all the frames in a sequence instead of a limited number of selected frames. The accuracy at every frame is computed using the methodology from the 2015 KITTI stereo benchmark:
we compare disparities (input disparities and disparity maps computed from the active reconstruction against the ground truth disparities derived from LIDAR),
and consider points whose delta disparity is greater than 3px and 5\% of the ground truth disparity as erroneous. The completeness of the input and fused disparity maps is computed as the percentage of ground truth points in the left camera's field of view which have a corresponding value in either
the input or the fused disparity maps. Conceptually, the accuracy and completeness scores are similar
to the precision and recall metrics used in information retrieval.

In order to separately evaluate the reconstruction accuracy of the background and the
dynamic objects in a manner which compares the fused map and the input disparity maps in a
fair way, we perform \emph{semantic-aware evaluation}.
To this end, we use the semantic segmentation results produced by the Multi-task Network
Cascades.
The input and fused depth maps are evaluated as follows:
(1) Ground truth points not associated with any potentially dynamic object are counted towards the \emph{static map} statistics. (2) Ground truth points associated with potentially dynamic objects which are being reconstructed are counted towards the \emph{dynamic object} statistics. (3) The remaining ground truth points which correspond to dynamic objects not undergoing reconstruction (e.g., bikes, pedestrians, or distant cars whose 3D motion cannot be computed reliably) are ignored. This method is, obviously, imperfect, as it relies on the computed semantic segmentation, which is not always fully reliable. Nevertheless, we have found it to work well in practice, allowing us to draw numerous insights about our system’s performance under various conditions, as will be described in detail below.

\begin{figure*}
  \centering
	\begin{subfigure}[t]{0.18\textwidth}
       \includegraphics[width=\linewidth]{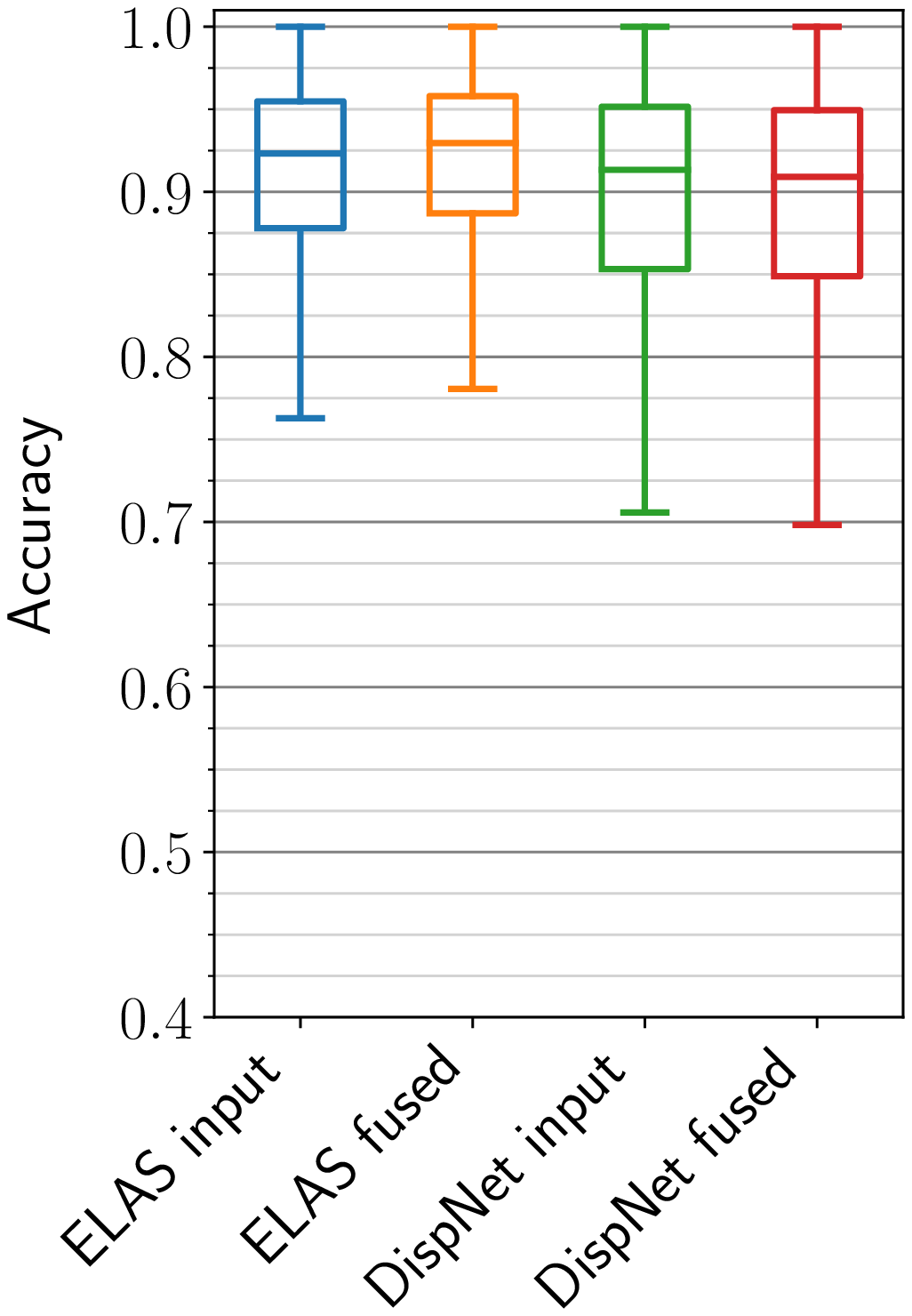}
       \caption{Reconstruction accuracy for the static background.}
    \end{subfigure}
    ~
	\begin{subfigure}[t]{0.18\textwidth}
      \includegraphics[width=\linewidth]{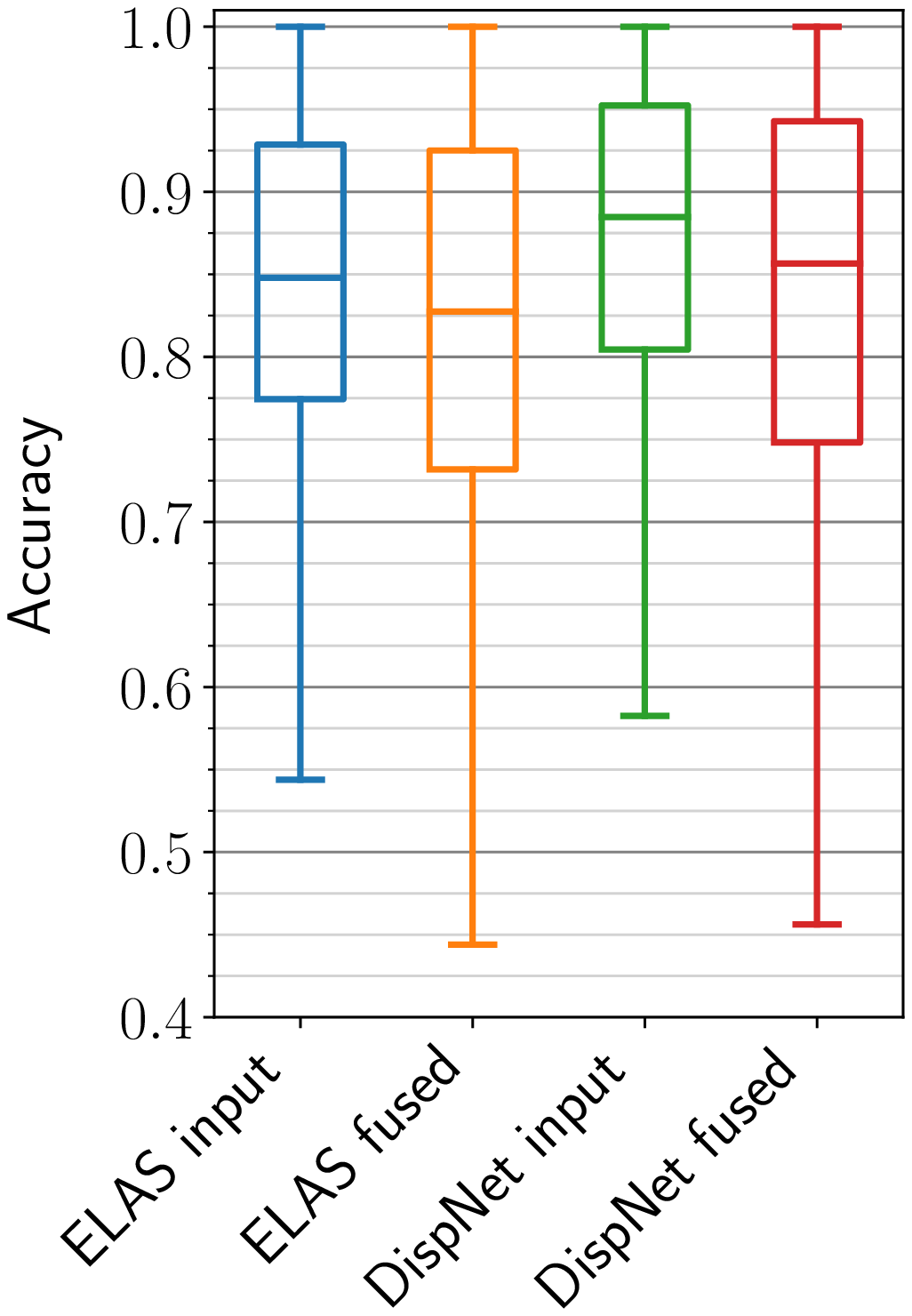}
      \caption{Reconstruction accuracy for the dynamic objects.}
    \end{subfigure}
    ~
	\begin{subfigure}[t]{0.18\textwidth}
       \includegraphics[width=\linewidth]{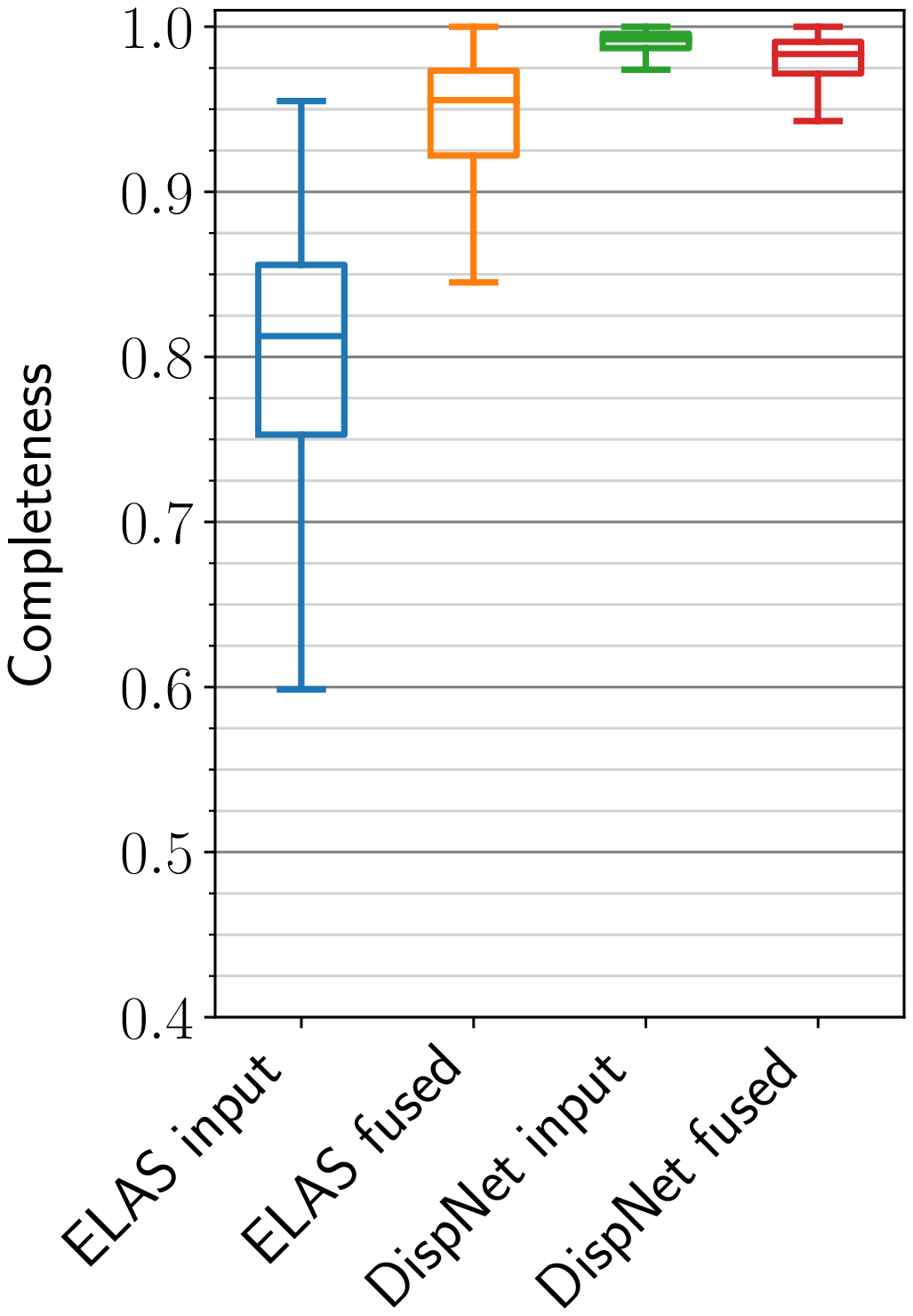}~
       \caption{Reconstruction com-pleteness for the static background.}
    \end{subfigure}
    ~
	\begin{subfigure}[t]{0.18\textwidth}
      \includegraphics[width=\linewidth]{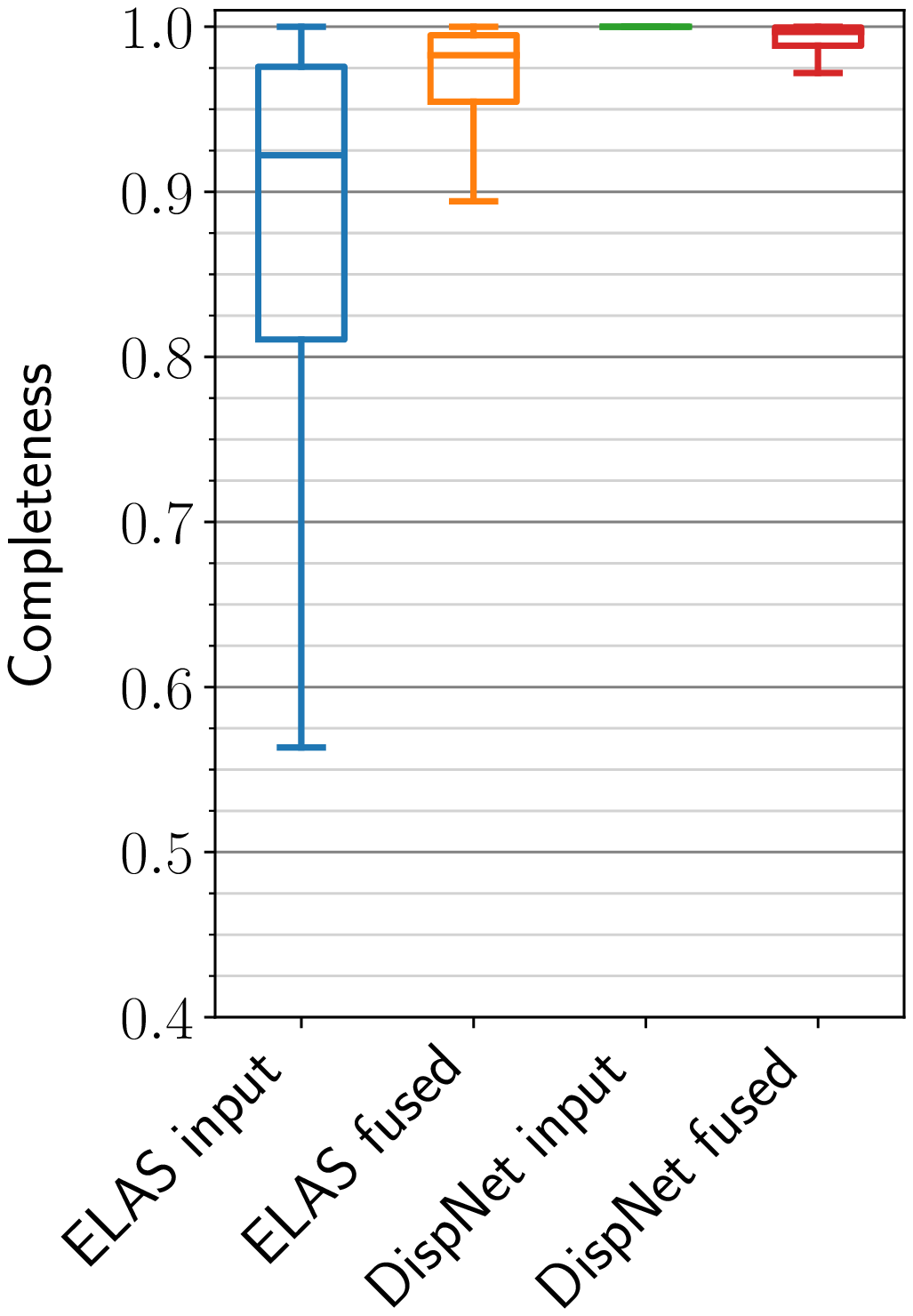}
      \caption{Reconstruction completeness for the dynamic objects.}
    \end{subfigure}
    \caption{Aggregate reconstruction accuracy and completeness on the KITTI odometry benchmark training sequences. The bottom and top edges of the boxes correspond to the first and third quartiles of the represented distribution, and the bottom and top whiskers stretch up to the lowest and highest data points still within 1.5 IQR (inter-quartile range) of the lower or upper quartile, following the Tukey convention.\label{fig:odo-both}}
\end{figure*}

\fref{fig:odo-both} presents the accuracy and completeness of the system's background
and object reconstructions, aggregated over all frames in the 11 KITTI odometry training sequences.
Several conclusions can be drawn from  the results.
\begin{enumerate*}
  \item Using ELAS depth maps leads to more accurate static maps
than DispNet, but less accurate object reconstructions. This is explained by
DispNet's superior robustness to challenges such as reflective and transparent
surfaces, which are much more widespread when reconstructing cars.
  \item The fusion process improves the reconstruction accuracy when using ELAS
depth maps, but not when using DispNet. This is due to the more conservative nature of the results produced by ELAS, which tend to have gaps but to be sharper overall.
The fusion process benefits from the sharpness while compensating for the occasional gaps.
  \item The reconstruction accuracy of the dynamic objects is usually not improved
by fusion when compared to the input maps, and the overall variance across frames
is much higher. This follows from the fact that vehicles are considerably more
challenging to reconstruct than, e.g., road surfaces, fences, and buildings,
due to their non-lambertian properties, as well as their independent motion.
  \item DispNet leads to much denser, i.e, more complete, reconstructions of both the static map and of the objects than ELAS, but this gap is reduced by volumetric fusion
over multiple frames. This follows from the fact that ELAS depth maps are sparser, but
more accurate, meaning that they benefit more from the fusion than DispNet.
\item The variance of the ELAS depth map completeness is high because it often produces very sparse results in challenging lighting conditions such as after exiting
a tunnel, before the automatic exposure of the camera manages to adapt. Nevertheless, the magnitude of this effect is reduced significantly by the fusion process.
\end{enumerate*}

Additionally, we compare the reconstruction accuracy of our system to two
baselines on the first ten sequences from the KITTI tracking benchmark.  The
first baseline is InfiniTAM itself, using its built-in ICP-based tracker.
Naturally, this baseline does not perform well, as InfiniTAM by itself is not
designed for outdoor environments and the fast camera motion encountered in the
KITTI dataset.  The next baseline is similar to the first, except that it uses
\texttt{libviso2} for robust visual odometry. However, just like the first, it
does not attempt to detect, remove, or reconstruct dynamic objects. The results
of this comparison are presented in~\tref{tab:baseline-comparison}.

As in the previous series of experiments, the ELAS-based results are more
accurate than their DispNet-based counterparts, but less complete.
Moreover, our system outperforms the robust VO InfiniTAM baseline in terms of
accuracy.
Despite its lower accuracy, the robust VO baseline does lead to slightly
higher completeness scores than dynamic fusion (our system).
This is explained by the fact that object instance removal is not perfect, and tends to sometimes also lead to small areas of the background being removed along with the objects. While this does not affect the quality of the instance reconstructions, with the additional background fragments being prime candidates for voxel garbage collection, it does explain the slightly lower scores of the dynamic fusion in terms of completeness.
These results showcase our method’s ability to improve the quality of the static maps by actively preventing dynamic objects from corrupting them.

\begin{table}
	\centering
   \caption{Aggregate accuracy and completeness scores on the first 10
   sequences from the KITTI tracking benchmark.\label{tab:baseline-comparison}}
   \begin{tabular}{@{}lrr@{}}
   \toprule
   \textbf{Method} & \textbf{Accuracy} & \textbf{Completeness} \\
   \midrule
   InfiniTAM + ELAS & tracking failure & tracking failure\\
   InfiniTAM + DispNet & tracking failure & tracking failure \\
   InfiniTAM + libviso2 + ELAS & 0.921 & 0.974\\
   InfiniTAM + libviso2 + DispNet & 0.875 & 0.984 \\
   Our system + ELAS & 0.923 & 0.961 \\
   Our system + DispNet & 0.879 & 0.973 \\
   \bottomrule
   \end{tabular}
\end{table}

\subsection{Effectiveness of fixed-lag map regularizer}
We evaluate the impact of the map regularization on the system's memory consumption and reconstruction accuracy. The methodology for evaluating the accuracy is the same as in the previous section. In order to also capture the effect of the voxel garbage collection, which is performed in lockstep with the reconstruction, but with a fixed delay of $k_{\text{minAge}}$ frames, we also add a delay to the evaluation. That is, at time $t$, the voxel garbage collection is processing the blocks visible at time $t - k_{\text{minAge}}$, and the evaluation is performed using the depth map and camera pose from $t - k_{\text{minAge}} - \tau$. The additional offset $\tau$ ensures that the map viewed by the camera at that time has been processed by the regularization. In our experiments, we set $\tau= k_{\text{minAge}}$.

Given the limitations of the ground truth, which is provided in the form of
per-frame LIDAR readings, we only evaluate the accuracy of the static
reconstructions under the effect of voxel garbage collection. The delayed
evaluation scheme described above prevents us from also evaluating dynamic
object reconstructions, as their positions are only known to the system while they
are being observed, and not $k_{\text{minAge}} + \tau$ frames in the past.

To this end, we use the same semantic-aware evaluation scheme defined previously
with the only difference being that we only evaluate the static map.
We perform our experiments on the first 1000 frames of KITTI odometry sequence
number 9, as it contains a small number of dynamic objects, while at the same
time being diverse in terms of encountered buildings and vegetation.
Even through we also use the regularization for the vehicle reconstructions,
they only represent a small fraction of the system's overall memory usage.
We therefore focus on evaluating the memory usage of the static map.

\begin{figure*}[htb]
\centering
\includegraphics[width=0.95\textwidth]{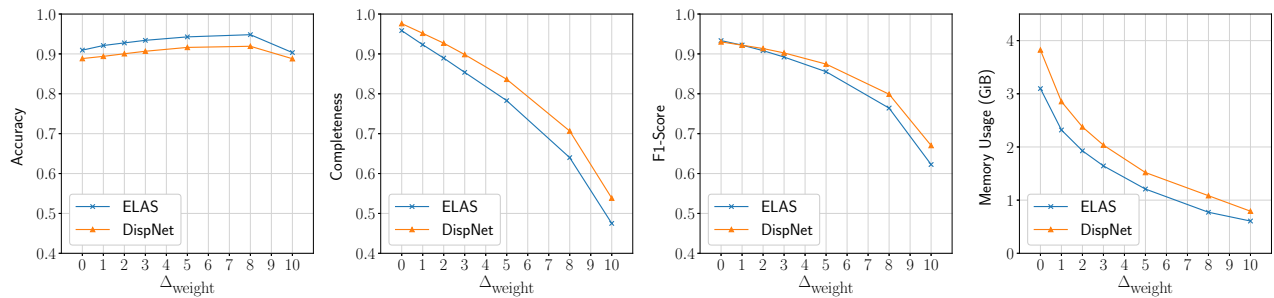}
\caption{Reconstruction accuracy, completeness, F1-score, and memory usage (in GiB) as functions of the
regularization strength, comparing its effect on reconstructions using ELAS and DispNet depth maps. Larger
values of $\Delta_{\text{weight}}$ correspond to more aggressive voxel garbage collection.\label{fig:regularization}}
\end{figure*}

\fref{fig:regularization} illustrates the accuracy and completeness of the
map reconstructions as a function of $\Delta_\text{weight}$, the noise threshold,
as well as the memory usage and an F1 score combining accuracy (A) and
completeness (C) as
\newcommand{\Acc}{\operatorname{A}}
\newcommand{\Com}{\operatorname{C}}
\newcommand{\cframe}{\text{frame}}
\begin{equation}
  \operatorname{F}_1(\cframe) = 2 \cdot \frac{\Acc(\cframe) \cdot
  \Com(\cframe)}{\Acc(\cframe) + \Com(\cframe)}.
\end{equation} The maps produced using DispNet are denser than those
based on ELAS, which is also reflected by the completeness of the
reconstruction. At the same time, as in the previous experiments, despite
being less accurate than DispNet according to the KITTI Stereo Benchmark~\cite{Menze2015}\footnote{\url{http://www.cvlibs.net/datasets/kitti/eval_scene_flow.php?benchmark=stereo}},
the depth maps produced by ELAS lead to more accurate, albeit less complete,
reconstructions. This trend is maintained even with increasing $\Delta_\text{weight}$.
Based on the memory usage plot from~\fref{fig:regularization}, it becomes clear that
the impact of the noise on the memory consumption of the reconstructions is strong.
Even light regularization using $\Delta_\text{weight} = 1$ or $2$ can already reduce
the memory footprint of a reconstruction by more than 30\%, with only small
costs in terms of discarded (useful) information.
The memory usage of the vanilla InfiniTAM system corresponds to $\Delta_\text{weight} = 0$.

For additional qualitative and quantitative results, we would like to direct
the reader to the video and the supplementary results\footnote{%
Available on the project's web page at \url{http://andreibarsan.github.io/dynslam}.}
accompanying this publication.

Our system is able to run at approximately 2.5Hz on a PC equipped with an
nVIDIA GeForce GTX 1080 GPU.

\section{CONCLUSIONS}
\label{sec:conclusions}
We presented a robust dense mapping algorithm for large-scale dynamic environments. 
Our algorithm employs a state-of-the-art deep convolution neural network for 
semantic segmentation, which separates the static background from the dynamic and 
potentially dynamic objects. 
The sparse scene flow is used to estimate both the camera egomotion, as well as the
individual motions of dynamic objects encountered in the environment.
The static background and rigid dynamic objects are then reconstructed separately. Real-world experiments show that our algorithm can output high-quality dense models of both the static background and the dynamic objects, 
which is important for high-level robotic decision making tasks such as path planning. 
Additionally, we proposed a map pruning technique based on voxel garbage collection
capable of improving the map accuracy, while substantially reducing memory consumption,
which is beneficial to resource-constrained mobile robots.

\bibliographystyle{plain}
\bibliography{root}

\end{document}